\makeatletter\def\graphicscache@inhibit{true}\makeatother

\documentclass[a4paper, 10pt, conference]{ieeeconf}      %

\IEEEoverridecommandlockouts                              %

\usepackage[utf8]{inputenc}
\usepackage{caption}
\usepackage{amssymb, amsthm, amsmath}
\usepackage{enumerate} 
\usepackage{float} 
\usepackage{graphicx}
\usepackage{graphicscache}
\usepackage{mathtools}
\usepackage{multirow}
\usepackage{siunitx}
\usepackage{subcaption}
\usepackage{tabularx}
\usepackage{todonotes}
\usepackage{tikz}
\usepackage{supertabular}

\newtheorem{lemma}{Lemma}

\newcommand{\figref}[1]{Fig.~\ref{#1}}
\newcommand{\tabref}[1]{Tab.~\ref{#1}}
\newcommand{\etal}{et al.}
\usepackage{xpatch}
\xpretocmd{\eqref}{Eq.~}{}{}

\makeatletter
\let\NAT@parse\undefined
\makeatother
\usepackage{hyperref}

\title{\LARGE \bf
 Rendering and Tracking the Directional TSDF:\\
 Modeling Surface Orientation for Coherent Maps
}

\author{Malte Splietker and Sven Behnke%
\thanks{This research has been supported by MBZIRC 2017 price money.%
  \newline All authors are with the Autonomous Intelligent Systems Group, University of Bonn, Germany. {\tt\small splietker@ais.uni-bonn.de}%
\newline 978-1-6654-1213-1/21/\$31.00 \textcopyright 2021 IEEE}%
}
 
\begin{document}

\maketitle
\thispagestyle{empty} %
\pagestyle{empty} %

\begin{tikzpicture}[remember picture,overlay]
\node[anchor=north west,align=left,font=\sffamily,yshift=-0.2cm] at (current page.north west) {%
  In: Proceedings of the 10th European Conference on Mobile Robots (ECMR) 2021
};

\end{tikzpicture}%

\begin{abstract}
  Dense real-time tracking and mapping from \mbox{RGB-D} images is an important tool for many robotic applications, such as navigation or grasping.
The recently presented Directional Truncated Signed Distance Function (DTSDF) is an augmentation of the regular TSDF and shows potential for more coherent maps and improved tracking performance. In this work, we present methods for rendering depth- and color maps from the DTSDF, making it a true drop-in replacement for the regular TSDF in established trackers. We evaluate and show, that our method increases re-usability of mapped scenes. Furthermore, we add color integration which notably improves color-correctness at adjacent surfaces.
\end{abstract}

\section{Introduction}

Since its first appearance in KinectFusion~\cite{Newcombe2011}, GPU accelerated TSDF algorithms have become a de-facto standard in scene reconstruction from depth images, leveraging inexpensive sensors and massive parallel processing on GPUs for good real-time performance.
By modeling the closest distance to the next surface with a signed distance function (SDF), geometry can be reconstructed by finding the zero-transition from positive (i.e. in front of the surface) to negative (i.e. behind the surface) values. In practice, this function is obtained by fusing measurements into a regular grid, the so called voxels, and interpolating between them.
The necessity to store both front- and backside of the surface implies, however, that there is a minimum thickness of objects that can be represented. Especially with thin objects, integration of new measurements might interfere with and contradict old data belonging to a different surface, leading to a corrupted model. We have explored this issue and introduced the concept of the Directional Truncated Signed Distance Function (DTSDF) in our previous work~\cite{Splietker2019}.
The DTSDF uses six TSDF volumes, one for each positive and negative coordinate axis, to store surface sections with different orientations. We proposed a method for fusing depth images into this data structure and for extracting meshes with a modified marching cubes algorithm. The latter is, however, not applicable for real-time tracking applications.

Instead, in this work we propose methods for rendering virtual camera views, which allows to use standard geometric ICP for real-time sensor motion tracking. Moreover, we introduce color integration into the DTSDF, which helps preserving color details of adjacent object surfaces, especially along sharp edges. With these additions, the DTSDF becomes a true replacement for the regular TSDF with only minor modifications to the fusion and rendering algorithms. We evaluate our method on well-known datasets and show, that the DTSDF has advantages in tracking certain types of sequences and is better at preserving the overall map for later reuse.

\begin{figure}[t]
  \centering
  \includegraphics[width=\linewidth]{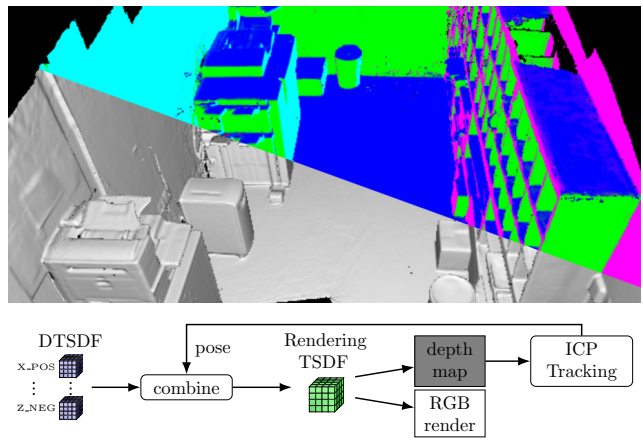}
  \caption{The top part shows a cut view of the reconstructed Stanford copyroom scene. The lower left half is the rendered depth, upper right shows the directions involved in rendering in different colors, mixed whenever multiple directions contributed. The bottom part shows the pipeline for rendering and tracking with the DTSDF.}
  \label{fig:iconic}
\end{figure}

\section{Related Work}
In 3D reconstruction and SLAM feature-based, sparse, and dense methods are distinguished. Our work belongs to the category of dense methods that describe closed surfaces, separate objects, and even free space~\cite{Oleynikova2017}.

Research focus in this field has shifted in recent years towards learned representations. Occupancy networks~\cite{Mescheder2019a} learn a binary descriptor describing the occupancy of space, i.e. whether a point lies inside an object or not. In DeepSDF by Park \etal~\cite{Park2019}, a representation is learned, which like our work allows querying the signed distance to the closest object for arbitrary points in space. Neural Radiance Fields (NeRF)~\cite{Mildenhall2020, Azinovic2021} use deep networks to regress density and color. While these approaches show impressive results, use less memory to store the model or even enable scene completion, they have some shortcomings. The limited model size results in a lack of detail for large scenes. Also training and inference times are still several orders of magnitude higher, making them unusable for real-time applications.
Hence, the classic TSDF fusion algorithms are still state-of-the-art in live mapping scenarios. 

Since its first occurrence, the TSDF fusion algorithm has seen widespread use cases and is mostly used without changes to the representation. There have been attempts to augment it, though. Dong \etal~\cite{Dong2018a} created a hybrid data structure, combining the TSDF with probabilistic surfels. Multiple overlapping TSDF sub-volumes are used in pose graphs for large-scale SLAM, enabling re-aligning parts of the map on loop closure detection for consistency~\cite{Henry2013, Millane2018}. This approach is similar to the DTSDF in that it maintains several overlapping representations, but does not fix the TSDF's inability to represent thin objects observed from opposite sides within the same volume. While this may not be an issue for some applications, object scans and walk-around type scenes profit from this capability.

Zhang \etal~\cite{Zhang2021} give an overview of current RGB-D SLAM algorithms. %
The typical method for localizing the sensor pose in the TSDF is frame-to-model geometric ICP~\cite{Newcombe2011}, where a back-projected point cloud from the previous position is used with the point-to-plane metric and Gauss Newton for minimizing the registration error. There are modified versions, such as the extended ICP tracker~\cite{Prisacariu2017}, which uses the Huber-norm and has advanced outlier detection. But the regular ICP is still most commonly used.

Photometric ICP instead uses a registration loss that is based on RGB values~\cite{Kerl2013}. This is helpful in geometrically ambiguous scenes, e.g. planar surfaces. Hybrid approaches combine both schemes~\cite{Henry2013, Whelan2015, Dai2017}.
Common for the ICP family of algorithms is, that they use depth maps and rendered images generated from the model for registering current input images.

The direct volume matching line of algorithms directly performs registration within the SDF. Point-to-SDF~\cite{Bylow2016, Canelhas2013} and SDF-to-SDF~\cite{Slavcheva2018} approaches can be distinguished.
Millane \etal~\cite{Millane2020} recently proposed a method for extracting and matching local features directly on the SDF.

Further hybrid approaches utilize other tracking sources. BundleFusion~\cite{Dai2017} is an advanced method that uses ICP error minimization and visual SIFT features in a global bundle adjustment, and then de- and reintegrates parts of scene to keep an overall representation consistent.

The goal of this work is to make DTSDF usable as replacement or supplement for the regular TSDF. To be able to profit from established tracking methods without further modifications we
\begin{itemize}
  \item introduce color fusion into the DTSDF,
  \item present an efficient method for generating rendered views of the DTSDF,
  \item use these rendered views to track sensor motion with the ICP algorithm.
\end{itemize}

\section{Fusion and Weights}
Formally, the signed distance function $\Phi:\mathbb{R}^3 \longrightarrow (d, w_d, \mathbf{c}, w_c)$ maps an arbitrary point in space to a tuple comprising signed distance to the closest surface $d$, distance weight $w_d$, RGB color $\mathbf{c}$ and color weight $w_c$, where the weights represent the confidence of the integrated information. As reconstruction only requires information close to the actual surface, the TSDF only maps points within a truncation band around the actual surface.
In practice, the TSDF is stored as a evenly-spaced grid of voxels and $\Phi$ is estimated by linear interpolation between the stored tuples.

The directional TSDF~\cite{Splietker2019} $\Phi_{\mathrm{dir}}(\mathbf{p}) = (\Phi^D(\mathbf{p}))_{D\in\textrm{Directions}}$ extends this representation by mapping a point to multiple signed distance functions -- one for each direction $\{X^+, X^-, Y^+, Y^-, Z^+, Z^-\}$ -- corresponding to the positive and negative coordinate axes $\mathbf{v} = \{(1, 0, 0)^\intercal, \cdots, (0, 0, -1)^\intercal \}$. Observed depth points are matched to those directions $D$ that fulfill $\arccos \langle \mathbf{n}, \mathbf{v}^D \rangle < \theta$ for depth normal $\mathbf{n}$ and angular threshold $\theta \in (\pi / 4, \pi /2]$.

\begin{figure}[h]
  \centering
  \includegraphics{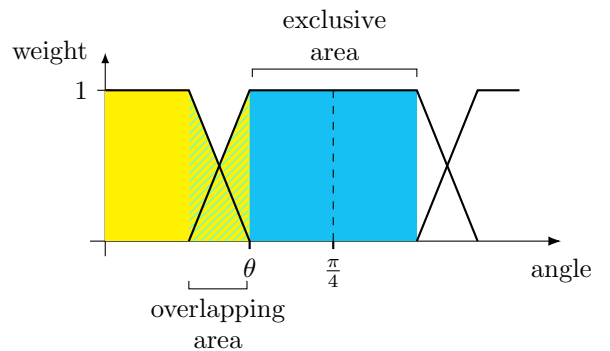}
  \caption{Direction weight function for angle between surface normal and direction vector. The angular threshold $\theta$ determines the width of exclusive (solid) and overlapping (hatched) areas for two neighboring direction (yellow, cyan).}
  \label{fig:direction_weight}
\end{figure}

Fusion is the process of integrating new observations into the voxels as weighted cumulative moving average, where weights denote the certainty of the added information. While there is no definite weighting scheme, most implementations use a combination of factors to compensate for measurement noise and for uncertainty by down-weighting voxels behind the surface \cite{Newcombe2011, Bylow2013, Nguyen2012}.
In addition to these factors, the fusion weight in~\cite{Splietker2019} includes a direction factor $w_{\textrm{dir}}^D = \left\langle \mathbf{n}, \mathbf{v}^D \right\rangle$ to linearly blend surfaces which fall into multiple directions over the whole span of $[0, \pi / 4]$. Then, all $w_{\textrm{dir}}^D$ are explicitly compared to threshold $\cos \theta$ and fused, if larger.
Instead, we propose the membership function
\begin{align}
  w_{\textrm{dir}}^D(\mathbf{n}) =
  \min\left(\max\left(
  \frac{1 - \arccos \langle \mathbf{n}, \mathbf{v}^D \rangle}{2 \theta - \frac{\pi}{4}}, 0
  \right), 1\right)
  \label{eq:direction_weight}
\end{align}
which has two advantages: firstly, it only blends overlapping parts up to $\theta$ with the full $[0, 1]$ range (cf.~\figref{fig:direction_weight}), whereas the old weight had an effective range of $[4 \theta/\pi, 1]$ and no exclusive area. Secondly, it makes explicit thresholding superfluous. 

Color is fused analogous to distances, but with a different weight. Again, there are different variants throughout literature. Dryanovski~\etal~\cite{Dryanovski2017} use the same constant weight for depth and color to save computation time.
Whelan \etal~\cite{Whelan2015} use angle between depth normal and view ray to downweight steep observation angles, which Bylow~\etal~\cite{Bylow2013} use in combination with the depth weight. This factor is also included in our depth weight.
We argue that using the depth fusion weight for colors is important, as the uncertainty is reflected in the choice of voxels associated with pixels. Our color weight is
\begin{align}
  w^{\mathrm{new}}_c = w_d^{\mathrm{new}} \left(1 - \min\left(1, \frac{||\mathbf{p} - \mathbf{x}||}{\tau}\right) \right),
  \label{eq:color_weight}
\end{align}
where $w_d^{\mathrm{new}}$ is the depth fusion weight, $\mathbf{p}$ the depth point and $\mathbf{x}$ the voxel position. The additional factor reduces the confidence for voxels further away from the surface, as multiple colors from various observations may blend together here and there is no equivalent to the point-to-plane metric, which mitigates this issue for depth fusion.

Free space, that is space between camera and observed surface, is not explicitly mapped to save memory. Nonetheless, due to noise, sensor error, or dynamic objects it can happen that spurious measurements are mapped in space, that is unoccupied in reality, and it is important to remove these artifacts.
When the computed distance~\eqref{eq:point-to-plane} is larger than the truncation range $\tau$, the voxel is located in free space and updated with a SDF value of 1 and a constant weight. No directional weight is used in this case, as the goal is to carve everything in free space. Special care has to be taken at depth discontinuities: carving can corrupt voxels of edges, because aliasing and small tracking inaccuracies associate the voxel with a more distant surface. Therefore, carving is only applied if there is no depth difference of more than $\tau$ in a radius of two pixels to the associated depth pixel.
To free up memory, voxels that are erroneously allocated but become free space by repeated carving can be recycled in an asynchronous process, as has been done in~\cite{Dong2018}.

The signed distance $\in [-1, 1]$ for a voxel position $\mathbf{x}$ is computed with the point-to-plane metric
\begin{align}
  d = \frac{1}{\tau} \langle \mathbf{p} - \mathbf{x}, \mathbf{n} \rangle
  \label{eq:point-to-plane}
\end{align}
for depth point $\mathbf{p}$ with normal $\mathbf{n}$ and truncation range $\tau$, which helps keeping the SDF consistent with varying observation angles as opposed to the point-to-point metric~\cite{Bylow2013}.

While in our previous work~\cite{Splietker2019} we explored ray casting similarly to Klingensmith \etal~\cite{Klingensmith2015} for fusing individual depth pixels along the view- or normal direction, this method often shows bad results with noisy real-world data. For tracking applications, voxel projection, like in the original KinectFusion, has proven more robust. During voxel projection fusion, all allocated voxels in the view frustum are projected into the current camera frame, associated with a depth (and color) pixel, and updated by these values.

\section{DTSDF Raycast Rendering}
Rendering real-time views of the model from arbitrary positions is useful for visualization and also tracking. 
Instead of developing specialized tracking methods for the DTSDF, our approach is to render a map of depth points and use known and tested ICP-based algorithms~\cite{Newcombe2011, Prisacariu2017, Kerl2013, Henry2013} to register input depth images.

The rendering process involves casting a ray per pixel of the virtual depth camera and extracting the iso-surface, i.e. the first transition from positive to negative SDF values. This involves probing the TSDF along the ray at regular intervals, until the distance turns negative and then multiple small steps, determined by the interpolated SDF value, to minimize the absolute distance value.
Similar to the meshing presented in~\cite{Splietker2019} the question is: how to combine up to six SDF values from partially overlapping directions?

\begin{figure}
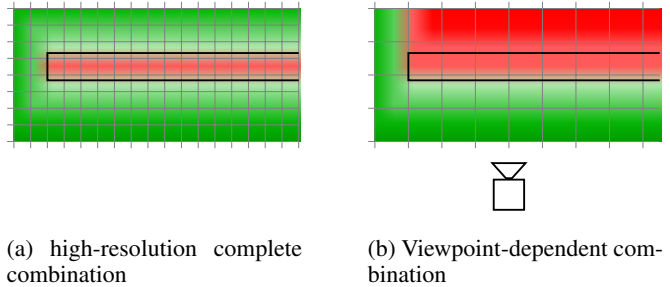

    \centering
  \begin{subfigure}[b]{0.45\linewidth}
    \includegraphics[width=\linewidth]{images/combination_3.pdf}
    \caption{high-resolution complete combination}
    \label{fig:positional_unambiguity_3}
  \end{subfigure}
  \hfill
  \begin{subfigure}[b]{0.45\linewidth}
    \includegraphics[width=\linewidth]{images/combination_4.pdf}
    \caption{Viewpoint-dependent combination}
    \label{fig:positional_unambiguity_4}
  \end{subfigure}
  \caption{Two variants of combining the DTSDF. The black outline represents the mapped object, green/red gradients correspond to the positive/negative SDF values and the grid denotes the voxel grid.}
  \label{fig:positional_unambiguity}
\end{figure}

By its mathematical definition, the signed distance function can represent any given object. In other words, the six directions could, given a fine enough resolution, be combined into a single, conflict-free TSDF. But in practice, the combination is not straight forward: overlapping free and occupied space from different volumes has to be combined in accordance with the orientation of mapped surfaces, while considering corner cases, real-world noise and imperfections.
Also, the practical use is limited. Ray-cast rendering relies on the width of the truncation range for finding zero-transitions, which for thin objects can be easily missed.
Instead, we made an important observation:
\begin{lemma}
  For a DTSDF and a fixed camera position, a combined conflict-free regular TSDF with the same voxel resolution can be computed.
  \label{lem:positional_unambiguity}
\end{lemma}
The basic intuition behind this lemma is, that surfaces the camera perceives from the backside are not relevant from a given position.
\figref{fig:positional_unambiguity} depicts a comparison of both variants. For the camera position in \figref{fig:positional_unambiguity_4}, the free space above the object is not required and the wider band of negative SDF values makes it easier to ray-cast.

Computing the true combination of directions is complicated and contains many corner cases. Instead we propose a simple weighting scheme as approximation.
For a point $\mathbf{x} \in \mathbb{R}^3$ and direction $D$, let $\mathbf{n}^D$ be the normalized SDF gradient $(\partial \Phi^D / \partial x,\partial \Phi^D / \partial y,\partial \Phi^D / \partial z)^\intercal$ at $\mathbf{x}$, $w_d^D$ the stored distance weight and $\mathbf{r}$ the normalized view ray from camera center to $\mathbf{x}$. Then the combination weight for direction $D$ is defined as
\begin{align}
  w_{\mathrm{comb}}^D = 
  w_{\textrm{dir}}^D(\mathbf{n}^D)
  \cdot
  \langle \mathbf{n}^D, -\mathbf{r} \rangle
  \cdot
  w_d^D.
  \label{eq:combine_weight}
\end{align}
The first factor in \eqref{eq:combine_weight} ensures that only gradients that actually comply with the direction they are stored in are used, with according weights to blend between directions. The second factor ensures that only directions with eligible surfaces are used, which is the main reason for using the DTSDF. The approximation is not perfect and certain constellations work only under the premise, that all direction's SDF weights are similar. On the other hand, it has shown to be more robust in practice than other, more sophisticated attempts we've tried.

These per-direction weights can be used to directly look up the combined SDF value at any point in space as weighted sum, but ray-casting becomes very slow, because many TSDF lookups and memory reads have to be performed, especially for the gradient computation. The massive parallel computation also results in many cache misses, so the algorithm becomes memory bound.
As suggested in Lemma~\ref{lem:positional_unambiguity}, a view-dependent combined TSDF can be pre-computed by calculating the combined SDF for every voxel in the view frustum. This combined TSDF can then be used like a regular TSDF, but only for ray-casting from the pose used during combination. As a bonus, this opens up yet another class of tracking algorithms: the direct volume matching type, that perform registration directly within the SDF~\cite{Bylow2016, Canelhas2013, Slavcheva2018}.

To always use the most recent observations, all voxels that received new information during fusion also need to be updated in the combined TSDF. But for static scenes this is not always necessary.
Instead, we use conditional combination. Only meeting one of the following criteria triggers an update of the combined TSDF:
\begin{align}
    \mathrm{framesSinceStart} &< 5,&&\textit{boot up}\label{eq:condition_1} \\ 
    \mathrm{framesSinceLastUpdate} &> 50,&&\textit{stale state}\label{eq:condition_2}\\
    \lVert \mathrm{pose} - \mathrm{lastPose}\rVert_{\textrm{translation}} &> \SI{0.05}{m},&&\textit{translation}\label{eq:condition_3} \\
    \lVert \mathrm{pose} - \mathrm{lastPose}\rVert_{\textrm{angle}} &> 0.05 \frac{\pi}{2}.&&\textit{rotation}\label{eq:condition_4}
\end{align}
The boot up condition~\eqref{eq:condition_1} ensures, that during the first frames where the map is still uncertain, always the most recent data is used for tracking. \eqref{eq:condition_2} enforces regular updates in case the camera does not move. Eq.~(\ref{eq:condition_3},~\ref{eq:condition_4}) are a relaxation of Lemma~\ref{lem:positional_unambiguity}, that states minor changes in the camera pose don't change the combined TSDF, similar to small-motion assumption on which the data association for ICP is based~\cite{Newcombe2011}.
We experimentally chose the thresholds relatively small, so as not to violate the underlying assumption. A more thorough investigation on the impact of these limits would be interesting. On the tested sequences, the update is triggered on average around every third frame.
By also selecting voxels slightly beyond the camera frustum ($\pm 1/8$ image size), motions of the camera will not leave the scope of the combined TSDF before triggering a recalculation. Voxels that receive data for the first time are always directly added to the combined TSDF.

To prevent empty voxels in the absence of gradients in all directions (e.g. at edges), the weights
\begin{align}
  w^D_{\mathrm{noGrad}} = w^D \cdot \langle \mathbf{v}^D, -\mathbf{r} \rangle
  \label{eq:combine_tsdf_weight}
\end{align}
are used instead.

\section{Implementation Details}
Our implementation is based on InfiniTAM~\cite{Kaehler2015}, with significant modifications. For optimized memory usage, the voxel hashing scheme introduced by Nießner \etal~\cite{Niesner2013} is used. Voxels are allocated in blocks of $8\times 8 \times 8$ only where required and a hash map is used for constant-time access. Among other changes, the implementation now uses the stdgpu library by Stotko~\cite{Stotko2019} to replace several components, especially the original hash map, which could not allocate blocks with colliding hash values within the same iteration.

Unlike the previous DTSDF implementation presented in~\cite{Splietker2019} where 6 separate TSDF volumes were used for the different directions, here only a single TSDF is utilized. The hash index is extended from $(x,y,z) \in \mathbb{Z}^3$ to $(x, y, z, D) \in \mathbb{Z}^3 \times \mathrm{Directions}$. %
This simplifies many functions and better utilizes the statically allocated memory on the GPU: For most scenes the DTSDF has an imbalance of direction-usage, which wastes a lot of memory in the old scheme. Resizing the volumes is an option, but requires additional overhead which can be simply avoided by the aforementioned modification.

The time for computing the rendering TSDF is crucial for real-time usage of our method. It can be significantly sped up by taking advantage of the lookup positions being only integer voxel positions. Hence, no trilinear interpolation is required and the gradient can be computed with just looking up the SDF values stored in the 6 neighboring voxels. We further managed to halve the time by pre-caching the TSDF lookups for all voxels of the same block in shared memory.
At the very most, for every direction there are the current block and its six neighboring blocks, so a total of 48 are looked up at the beginning.

As a proof of concept, we use the pipeline depicted in \figref{fig:iconic} with the default geometric ICP tracker.

\section{Evaluation}
\begin{figure}
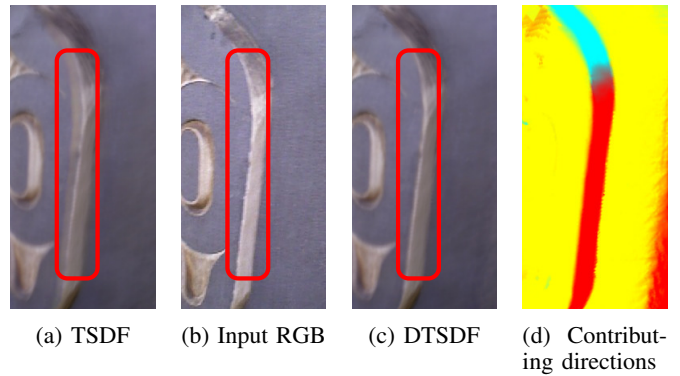

  \centering
  \subcaptionbox{TSDF\label{subfig:color_bleed_TSDF}}[.22\linewidth]
  {
    \centering
    \begin{tikzpicture}
      \node[anchor=south west,inner sep=0] (image) at (0,0) {\includegraphics[width=.107\textwidth,trim={230px 50 230 50},clip]{images/totempole_rgb_def_1428.png}};
      \begin{scope}[x={(image.south east)},y={(image.north west)}]
        \draw[red,ultra thick,rounded corners] (0.32,0.10) rectangle (0.60,0.85);
      \end{scope}
    \end{tikzpicture}
  }
  \hfill
  \subcaptionbox{Input RGB\label{subfig:color_bleed_rgb}}[.22\linewidth]
  {
    \centering
    \begin{tikzpicture}
      \node[anchor=south west,inner sep=0] (image) at (0,0) {\includegraphics[width=.107\textwidth,trim={230px 50 230 50},clip]{images/totempole_input_rgb_1428.png}};
      \begin{scope}[x={(image.south east)},y={(image.north west)}]
        \draw[red,ultra thick,rounded corners] (0.32,0.10) rectangle (0.60,0.85);
      \end{scope}
    \end{tikzpicture}
  }
  \hfill
  \subcaptionbox{DTSDF\label{subfig:color_bleed_DTSDF}}[.22\linewidth]
  {
    \centering
    \begin{tikzpicture}
      \node[anchor=south west,inner sep=0] (image) at (0,0) {\includegraphics[width=.107\textwidth,trim={230px 50 230 50},clip]{images/totempole_rgb_dir_1428.png}};
      \begin{scope}[x={(image.south east)},y={(image.north west)}]
        \draw[red,ultra thick,rounded corners] (0.32,0.10) rectangle (0.60,0.85);
      \end{scope}
    \end{tikzpicture}
  }
  \hfill
  \subcaptionbox{Contributing directions\label{subfig:color_bleed_directions}}[.22\linewidth]
  {
    \centering
    \begin{tikzpicture}
      \node[anchor=south west,inner sep=0] (image) at (0,0) {\includegraphics[width=.107\textwidth,trim={230px 50 230 50},clip]{images/totempole_directions_1428.png}};
    \end{tikzpicture}

  }
  \caption{Color bleeding effect on Stanford totempole sequence}
  \label{fig:color_bleed}
\end{figure}
\figref{fig:color_bleed} clearly shows the DTSDF's advantage in color separation. While in the regular TSDF (\figref{subfig:color_bleed_TSDF}) the colors blend because of fusion from two surfaces into the same voxels, the DTSDF retains different colors across edges (\figref{subfig:color_bleed_DTSDF}). \figref{subfig:color_bleed_directions} shows which directions contribute to which rendered pixel.

The datasets used in our evaluation are the Stanford 3D Scene Data (totempole, etc.)~\cite{Zhou2013}, ICL NUIM~\cite{Handa2014}, Zhou~\cite{Choi2015} and the TUM RGB-D benchmark \cite{Sturm2012}.
The sequences ICL lr0 and ICL office kt1 are skipped, because they contain a geometrically ambiguous segment which is not trackable with geometric ICP.

\subsection{ICP Tracking}
For comparing the tracking performance, we evaluate the regular TSDF (marked state-of-the-art, SoTA) against the DTSDF by running scenes from the aforementioned datasets and comparing the tracking results against the provided ground-truth trajectory using the relative pose error (RPE) with a window size of 30 frames (\SI{1}{s}). Note that this study does not try to compare to complete SLAM algorithm with loop closure detection and correction, but showcases the performance of the DTSDF relative to the regular TSDF as an enhanced data structure. All settings are equal across both modes and the tracker uses the default geometric ICP algorithm.

As expected, tracking does benefit from the DTSDF in scenes, where the camera observes thin structure from different angles. Otherwise, there is no significant difference.

The first test on artificially generated sequences from the ICL NUIM and Zhou datasets is reported in \tabref{tab:tracking_ICL} for different voxel sizes. Note that the noise-augmented sequences are being used.
The tracking performance is similar for most sequences, which is likely due to the mapped environments, which are convex rooms where the regular TSDF does not display its issues.
The Zhou office sequences, a scene of cluttered office desks scanned from different directions, provides an environment where the DTSDF actually has an advantage, which is reflected in the RPE.
\tabref{tab:tracking_models} does the same comparison on the turntable-like dataset used in the original DTSDF paper~\cite{Splietker2019}. In those sequences the camera orbits around a center point and only the model is visible, which is challenging to track due to the details and high percentage of thin structure wrt. to the whole scene. The RPE distinctly shows the strength of the DTSDF.
\tabref{tab:tracking_tum} shows the results with real-world scans from the TUM dataset, with very similar results. Large planar surfaces with sharp corners (structure notex sequences) seem to benefit from the DTSDF. Here, the measurements have to be considered with care, as the underlying ground-truth is not perfect. 

\newcolumntype{R}[1]{>{\raggedleft\arraybackslash}p{#1}}
\begin{table}[h]
  \caption{Tracking RPE in \SI{}{mm}, mean memory usage, and per-frame runtime of synthetic ICL NUIM~\cite{Handa2014} and Zhou~\cite{Choi2015} sequences for different voxel sizes.}
  \label{tab:tracking_ICL}
  \center
  \begin{tabular}{p{16mm}|R{6mm}R{8mm}|R{6mm}R{8mm}|R{6mm}R{8mm}}
    Voxel size & \multicolumn{2}{c|}{5}& \multicolumn{2}{c|}{10}& \multicolumn{2}{c}{20} \\
    & SoTA & DTSDF & SoTA & DTSDF & SoTA & DTSDF\\
    \hline
    lr kt1n & 4.8 & 4.9 & 4.9 & 4.8 & 5.3 & 4.7 \\
    lr kt2n & 15.2 & 15.2 & 15.1 & 15.0 & 14.6 & 14.8 \\
    lr kt3n & 13.1 & 16.5 & 21.9 & 19.6 & 55.8 & 26.2 \\
    office kt0n & 11.8 & 11.4 & 11.2 & 11.3 & 11.1 & 11.2 \\
    office kt2n & 17.1 & 17.4 & 16.0 & 16.1 & 16.4 & 16.2 \\
    office kt3n & 185.4 & 191.6 & 299.1 & 287.1 & 640.2 & 326.2 \\
    Zhou lr1 & 2.9 & 2.9 & 2.9 & 3.0 & 4.1 & 4.0 \\
    Zhou lr2 & 2.5 & 2.4 & 2.6 & 2.8 & 4.0 & 3.6 \\
    Zhou office1 & 1.9 & 1.7 & 4.0 & 1.8 & 10.3 & 3.0 \\
    Zhou office2 & 2.9 & 2.4 & 5.8 & 3.3 & 12.5 & 4.3 \\
    \hline
    $\varnothing$ time [\SI{}{ms}] 
    & 10.2 & 17.1 & 6.7 & 8.8 & 5.6 & 6.4 \\
    \hline
    $\varnothing$ mem [\SI{}{MB}] 
    & 2263 & 3117 & 462 & 769 & 85 & 169
  \end{tabular}
\end{table}

\begin{table}[h]
  \caption{Tracking RPE in \SI{}{mm}, mean memory usage, and per-frame runtime of synthetic sequences rendered from Stanford 3D models~\cite{Splietker2019} for different voxel sizes.}
  \label{tab:tracking_models}
  \center
  \begin{tabular}{p{16mm}|R{6mm}R{8mm}|R{6mm}R{8mm}|R{6mm}R{8mm}}
    Voxel size & \multicolumn{2}{c|}{5}& \multicolumn{2}{c|}{10}& \multicolumn{2}{c}{20} \\
    & SoTA & DTSDF & SoTA & DTSDF & SoTA & DTSDF\\
    \hline
    armadillo & 6.0 & 8.0 & 8.1 & 7.1 & 18.6 & 13.7 \\
    bunny & 3.7 & 3.0 & 4.2 & 3.8 & 9.9 & 7.5 \\
    dragon & 11.7 & 6.4 & 6.9 & 6.2 & 26.1 & 13.8 \\
    turbine blade & 14.4 & 11.0 & 32.0 & 38.7 & 93.0 & 32.7 \\
    \hline
    $\varnothing$ time [\SI{}{ms}] 
    & 4.8 & 5.1 & 4.4 & 4.7 & 4.2 & 4.5 \\
    \hline
    $\varnothing$ mem [\SI{}{MB}] 
    & 18 & 39 & 4 & 12 & 1 & 4
  \end{tabular}
\end{table}

\begin{table}[h!]
  \caption{Tracking RPE in \SI{}{mm}, mean memory usage, and per-frame runtime of TUM sequences~\cite{Sturm2012} for different voxel sizes.}
  \label{tab:tracking_tum}
  \center
  \begin{tabular}{p{16mm}|R{6mm}R{8mm}|R{6mm}R{8mm}|R{6mm}R{8mm}}
    Voxel size & \multicolumn{2}{c|}{5}& \multicolumn{2}{c|}{10}& \multicolumn{2}{c}{20} \\
    & SoTA & DTSDF & SoTA & DTSDF & SoTA & DTSDF\\
    \hline
    desk1 & 62.4 & 60.9 & 55.2 & 56.1 & 49.2 & 50.3 \\
    long office & 23.3 & 23.9 & 22.6 & 22.2 & 24.6 & 23.7 \\
    \multirow{2}{*}{\shortstack[l]{structure\\notex far}} & 31.9 & 26.8 & 33.8 & 33.8 & 17.8 & 18.9 \\
    & & & & \\
    \multirow{2}{*}{\shortstack[l]{structure\\notex near}} & 11.1 & 10.9 & 11.4 & 10.1 & 10.8 & 9.7 \\
    & & & & \\
    \hline
    $\varnothing$ time [\SI{}{ms}] 
    & 7.5 & 11.9 & 5.9 & 7.2 & 5.4 & 6.0 \\
    \hline
    $\varnothing$ mem [\SI{}{MB}] 
    & 681 & 1368 & 132 & 317 & 29 & 82
  \end{tabular}
\end{table}

\subsection{Map Reusability}
\begin{figure}[h]
  \centering
  \includegraphics[width=\linewidth]{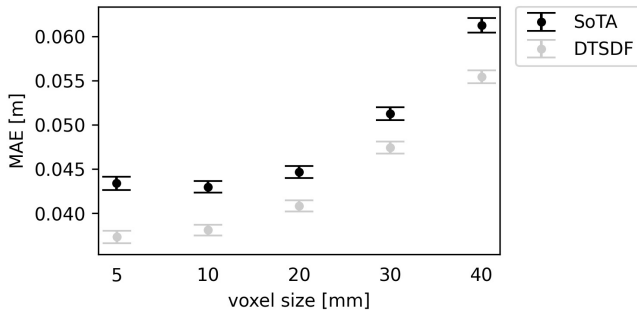}
  \caption{Post-fusion MAE (dot) and 95\% confidence intervals (bars) on example dataset TUM fr3 long office.}
  \label{fig:icp_error}
\end{figure}

Overall, the tracking results show, that the DTSDF generally only has an advantage in sequences, where the camera observes structure from different sides. In those sequences, the performance is significantly better, especially with increasing voxel size, as the problems of the regular TSDF increase as well.
In all other sequences the performance is very similar, which is also due to the fact that the fusion process makes the locally visible area compliant: given enough observations from the current viewpoint, all conflicts in the representation will be evened out due to the running average (unless the conflicting side has been observed a long time, resulting in a high weight).
In many cases this does not become apparent during tracking, but for the reusability of the overwritten parts of the completed map it is important to test these effects. To this end, we propose the post-fusion per-frame error: after completing fusion of the entire sequence, for every estimated pose, a depth map is rendered again and compared to the corresponding input depth image by computing the pixel-wise mean absolute error (MAE).
\figref{fig:icp_error} shows on the example of the \textit{fr3 long office} sequence that although the tracking performance is very similar, the DTSDF is better at retaining the map.

\begin{figure}[h]
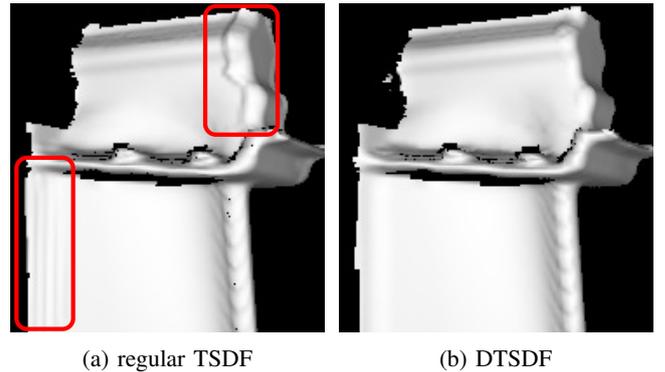

  \subcaptionbox{regular TSDF\label{subfig:quality_turbine_regular}}[.49\linewidth]
  {
    \centering
    \begin{tikzpicture}
      \node[anchor=south west,inner sep=0] (image) at (0,0) {\includegraphics[width=.232\textwidth,trim={230px 190 230 100},clip]{images/turbine_def_0680.png}};
      \begin{scope}[x={(image.south east)},y={(image.north west)}]
        \draw[red,ultra thick,rounded corners] (0.02,0.01) rectangle (0.20,0.53);
        \draw[red,ultra thick,rounded corners] (0.62,0.60) rectangle (0.85,0.99);
      \end{scope}
    \end{tikzpicture}
  }
  \subcaptionbox{DTSDF\label{subfig:quality_turbine_DTSDF}}[.49\linewidth]
  {
    \centering
    \begin{tikzpicture}
      \node[anchor=south west,inner sep=0] (image) at (0,0) {\includegraphics[width=.232\textwidth,trim={230px 190 230 100},clip]{images/turbine_dir_0680.png}};
    \end{tikzpicture}
  }
  \hfill
  \caption{Qualitative comparison of regular TSDF and DTSDF on turntable-style sequences. The lower left rectangle highlights artifacts from data fused from the backside. The upper right rectangle shows artifacts resulting from fusion conflicts between right- and front side.}
  \label{fig:quality_turbine}
\end{figure}

\tabref{tab:mae_selection} lists the post-fusion MAE of a selection of sequences (a complete list can be found in the Appendix~\tabref{tab:mae_datasets}). One can observe that in most cases there is not too much difference, especially in concave rooms like the ICL sequences. The effect usually only affects small parts of the model, like a corner or a computer monitor. Consequently, for the mean error over the whole image the effect is not that significant, but visible nonetheless.
Object-scanning type sequence with the camera orbiting around objects generally seem to profit from the DTSDF more (c.f.~\figref{fig:quality_turbine}). Decreasing the voxel size certainly does mitigate some of these issues for the SoTA, but ultimately the effect highlighted by the lower left rectangle remains for thin surfaces. Also, as \tabref{tab:tracking_ICL},~\ref{tab:tracking_tum} show, halving the voxel size instead of using the DTSDF will require more memory and computation time.

\begin{table}[ht]
  \caption{MAE (in \SI{}{\mm}) for different voxel sizes and datasets.}
  \label{tab:mae_selection}
  \center
  \begin{tabular}{p{19mm}l|p{5.5mm}p{5.5mm}p{5.5mm}p{5.5mm}p{5.5mm}}
    & & \multicolumn{5}{c}{Voxel size [\SI{}{\mm}]} \\
    Dataset & mode & 5 & 10 & 20 & 30 & 40 \\
    \hline
    \multirow{2}{*}{\shortstack[l]{SUN copyroom}} & SoTA & 56.4 & 36.5 & 53.3 & 65.9 & 92.8 \\
    & DTSDF & 51.2 & 37.1 & 35.5 & 47.0 & 62.2 \\
    \hline
    \multirow{2}{*}{\shortstack[l]{SUN stonewall}} & SoTA & 78.4 & 33.3 & 35.7 & 45.6 & 216.8 \\
    & DTSDF & 32.5 & 65.8 & 34.8 & 38.1 & 83.6 \\
    \hline
    \multirow{2}{*}{\shortstack[l]{ICL lr kt1n}} & SoTA & 55.3 & 77.8 & 128.3 & 183.7 & 207.1 \\
    & DTSDF & 46.9 & 58.4 & 103.1 & 111.6 & 115.4 \\
    \hline
    \multirow{2}{*}{\shortstack[l]{ICL office kt3n}} & SoTA & 218.3 & 153.6 & 85.1 & 104.6 & 95.3 \\
    & DTSDF & 40.0 & 252.4 & 121.8 & 141.8 & 145.3 \\
    \hline
    \multirow{2}{*}{\shortstack[l]{Zhou office2}} & SoTA & 25.9 & 29.0 & 34.3 & 84.3 & 64.6 \\
    & DTSDF & 25.8 & 28.1 & 31.8 & 37.1 & 44.6 \\
    \hline
    \multirow{2}{*}{\shortstack[l]{turbine blade}} & SoTA & 2.5 & 5.1 & 13.8 & 37.0 & 42.3 \\
    & DTSDF & 2.3 & 3.6 & 7.0 & 11.9 & 23.9 \\
    \hline
    \multirow{2}{*}{\shortstack[l]{TUM fr1 desk1}} & SoTA & 62.2 & 49.7 & 49.0 & 51.9 & 56.1 \\
    & DTSDF & 51.7 & 47.9 & 48.1 & 50.1 & 54.3 \\
    \hline
    \multirow{2}{*}{\shortstack[l]{TUM fr3 long\\office}} & SoTA & 43.4 & 43.0 & 44.7 & 51.2 & 61.3 \\
    & DTSDF & 37.3 & 38.1 & 40.8 & 47.4 & 55.4 \\
  \end{tabular}
\end{table}

\subsection{Runtime and Memory Consumption}
\begin{figure}[ht]
  \centering
  \begin{tikzpicture}
    \node at (0, 0) {\includegraphics[width=\linewidth]{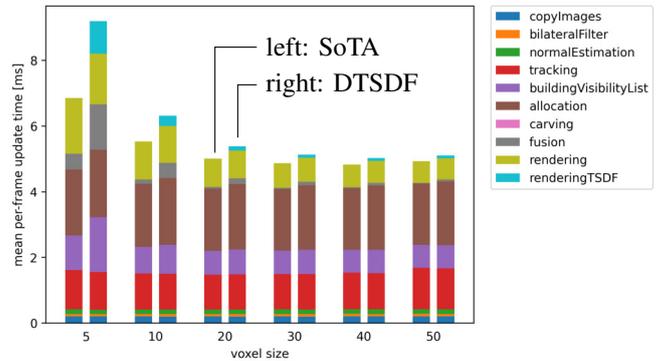}};
    \draw (-1.55, 0.4) -- (-1.55, 1.8) -- (-1.0, 1.8) node[anchor=west, align=left] {left: SoTA};
    \draw (-1.25, 0.6) -- (-1.25, 1.3) -- (-1.0, 1.3) node[anchor=west, align=left] {right: DTSDF};
  \end{tikzpicture}
  \caption{Mean per-frame update time comparison between state-of-the-art and DTSDF on SUN totempole sequence for different voxels sizes. Conditional combination is activated.}
  \label{fig:runtime_totempole}
\end{figure}

All experiments were performed on an Intel i7-8700K with 3.70GHz and a GeForce RTX 3090. The CPU part of the code runs entirely on a single core. By reducing the changes to rendering the DTSDF while keeping the rest of the pipeline original, the runtime only differs in allocation, fusion and rendering. \figref{fig:runtime_totempole} breaks down and compares runtimes for different voxels sizes. One can observe that the additional overhead is quite small. Note that for this example conditional combination was activated, and a rendering TSDF was computed for 35-40\% of the frames with the conditions specified in~\eqref{eq:condition_1}-(\ref{eq:condition_4}).

\begin{figure}
  \centering
  \includegraphics[width=\linewidth]{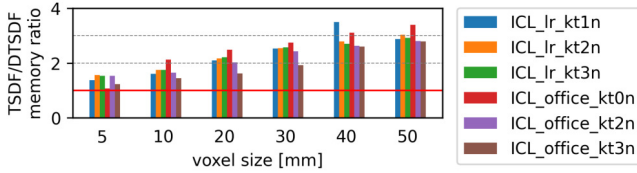}
  \caption{Ratio of allocated memory for DTSDF w.r.t. regular TSDF for different voxel sizes and scenes form the ICL NUIM dataset~\cite{Handa2014}.}
  \label{fig:allocation_ratio_ICL}
\end{figure}

As expected, the memory usage of the DTSDF is higher, as surfaces can overlap in up to three directions. In~\figref{fig:allocation_ratio_ICL} the ratio of additional memory required by the DTSDF w.r.t. the regular TSDF is displayed for ICL NUIM scenes.
For smaller voxels, the amount for extra memory is quite small. With increasing voxel size, the ratio increases as blocks are allocated in chunks of $8\times8\times8$ voxels and more surfaces with different orientations fall into the same block, though the actual number of blocks is significantly smaller.

\begin{figure}
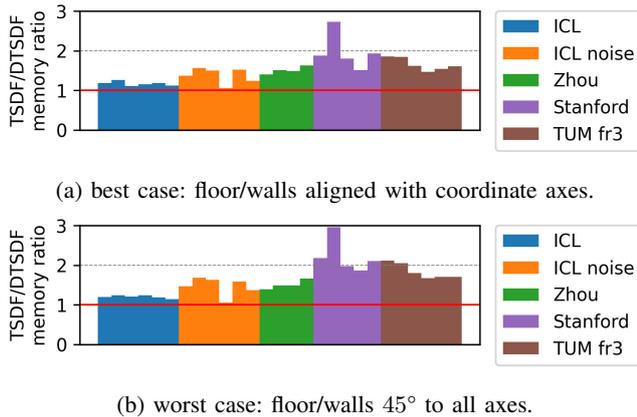

  \centering
  \begin{subfigure}[b]{\linewidth}
    \includegraphics[width=\linewidth]{images/allocation_ratio_best_case.png}
    \caption{best case: floor/walls aligned with coordinate axes.}
    \label{fig:allocation_best_case}
  \end{subfigure}
  \begin{subfigure}[b]{\linewidth}
    \includegraphics[width=\linewidth]{images/allocation_ratio_worst_case.png}
    \caption{worst case: floor/walls \ang{45} to all axes.}
    \label{fig:allocation_worst_case}
  \end{subfigure}
  \caption{Ratio of allocated memory for DTSDF wrt. regular TSDF (red line) for various sequences of different datasets and \SI{5}{mm} voxel size.}
  \label{fig:allocation_ratio_datasets}
\end{figure}

\figref{fig:allocation_ratio_datasets} plots the memory ratio for various sequences of the datasets fused with \SI{5}{mm} voxel size. It is also noticeable, that synthetic datasets (ICL, Zhou) use less memory than real ones. This is probably noise-related, as the depth-noise augmented ICL sequences also have a higher ratio, which suggests that with a more conservative allocation scheme memory can be saved. At the moment even for stray measurements blocks are allocated, as long as they have a valid normal.
As surfaces can be fused into up to three directions, determined by \eqref{eq:direction_weight}, an interesting question is how the alignment of the map coordinate frame to the scene affects memory usage.
For this, we pre-computed initial poses for each sequence by identifying the largest planes of the sequence.
In the best case scenario (\figref{fig:allocation_best_case}), the coordinate axes are parallel; in the worst case (\figref{fig:allocation_worst_case}), tilted \ang{45} to the identified planes. Noticeably, the alignment does have an effect, but not very significant. This is, however, highly dependent on the scenes and voxel sizes, because it induces only a one-time cost. Automating this process at the initialization phase of mapping is recommended.

Overall for the majority of scenes the DTSDF requires around 1.5 to 2 times as much memory as the regular TSDF.

\section{Conclusions}

In this work, we introduced the tools to use the DTSDF as a drop-in replacement for regular TSDF for mapping, tracking, and visualization applications. The ability to simply extract a regular TSDF for a given pose enables using it for a variety of tasks and with many algorithms that have been developed over the years.

We have shown that the DTSDF has advantages over the state-of-the-art for certain types sequences, like object scans, in both quality and quantity. Moreover, with the post-fusion MAE metric we shown that, while the regular TSDF is usable for local maps, reusability and revisiting of mapped places becomes problematic, if conflicting information from different surfaces corrupts the model. With reasonable memory- and computation overhead, better results and a more consistent map can be obtained by the proposed DTSDF method.

\section*{APPENDIX}

\vspace{-4mm}
\begin{table}[H]
\caption{MAE (in \SI{}{\mm}) of state-of-the-art and DTSDF compared for different voxel sizes and datasets.}
\label{tab:mae_datasets}
\footnotesize
\begin{tabular}{p{19mm}l|p{5.5mm}p{5.5mm}p{5.5mm}p{5.5mm}p{5.5mm}}
  & & \multicolumn{5}{c}{Voxel Size [\SI{}{\mm}]} \\
  dataset & mode & 5 & 10 & 20 & 30 & 40 \\
  \hline
  \multirow{2}{*}{\shortstack[l]{SUN burghers}} & SoTA & 155.5 & 63.1 & 51.8 & 54.5 & 61.3 \\
  & DTSDF & -- & 59.4 & 52.2 & 52.9 & 57.4 \\
  \hline
  \multirow{2}{*}{\shortstack[l]{SUN copyroom}} & SoTA & 56.4 & 36.5 & 53.3 & 65.9 & 92.8 \\
  & DTSDF & 51.2 & 37.1 & 35.5 & 47.0 & 62.2 \\
  \hline
  \multirow{2}{*}{\shortstack[l]{SUN cactusgarden}} & SoTA & 61.0 & 57.8 & 62.2 & 72.2 & 83.7 \\
  & DTSDF & 56.5 & 57.7 & 65.4 & 74.4 & 82.7 \\
  \hline
  \multirow{2}{*}{\shortstack[l]{SUN lounge}} & SoTA & 61.3 & 61.1 & 63.2 & 67.1 & 72.6 \\
  & DTSDF & 59.2 & 60.8 & 62.9 & 65.7 & 69.6 \\
  \hline
  \multirow{2}{*}{\shortstack[l]{SUN stonewall}} & SoTA & 78.4 & 33.3 & 35.7 & 45.6 & 216.8 \\
  & DTSDF & 32.5 & 65.8 & 34.8 & 38.1 & 83.6 \\
  \hline
  \multirow{2}{*}{\shortstack[l]{SUN totempole}} & SoTA & 25.0 & 29.9 & 31.3 & 34.4 & 39.3 \\
  & DTSDF & 24.9 & 29.7 & 30.7 & 33.0 & 36.4 \\
  \hline
  \multirow{2}{*}{\shortstack[l]{ICL lr kt1}} & SoTA & 17.8 & 18.2 & 17.7 & 17.6 & 18.0 \\
  & DTSDF & 17.8 & 18.1 & 17.7 & 17.6 & 17.8 \\
  \hline
  \multirow{2}{*}{\shortstack[l]{ICL lr kt2}} & SoTA & 24.1 & 25.5 & 25.6 & 26.6 & 28.8 \\
  & DTSDF & 24.1 & 25.5 & 25.6 & 26.5 & 28.7 \\
  \hline
  \multirow{2}{*}{\shortstack[l]{ICL lr kt3}} & SoTA & 79.8 & 67.7 & 22.6 & 17.7 & 65.9 \\
  & DTSDF & 123.9 & 69.8 & 67.5 & 74.8 & 19.8 \\
  \hline
  \multirow{2}{*}{\shortstack[l]{ICL lr kt1n}} & SoTA & 55.3 & 77.8 & 128.3 & 183.7 & 207.1 \\
  & DTSDF & 46.9 & 58.4 & 103.1 & 111.6 & 115.4 \\
  \hline
  \multirow{2}{*}{\shortstack[l]{ICL lr kt2n}} & SoTA & 61.1 & 74.2 & 94.4 & 108.3 & 164.2 \\
  & DTSDF & 49.3 & 71.2 & 100.2 & 112.4 & 142.9 \\
  \hline
  \multirow{2}{*}{\shortstack[l]{ICL lr kt3n}} & SoTA & 35.2 & 42.9 & 119.8 & 112.2 & 150.7 \\
  & DTSDF & 38.2 & 41.7 & 53.3 & 68.7 & 106.0 \\
  \hline
  \multirow{2}{*}{\shortstack[l]{ICL office kt0}} & SoTA & 20.9 & 24.6 & 26.4 & 27.7 & 28.1 \\
  & DTSDF & 20.9 & 24.7 & 26.5 & 27.6 & 28.1 \\
  \hline
  \multirow{2}{*}{\shortstack[l]{ICL office kt2}} & SoTA & 21.1 & 23.7 & 24.4 & 24.5 & 25.4 \\
  & DTSDF & 21.1 & 23.8 & 62.6 & 24.6 & 25.3 \\
  \hline
  \multirow{2}{*}{\shortstack[l]{ICL office kt3}} & SoTA & 14.9 & 15.8 & 16.0 & 16.1 & 16.7 \\
  & DTSDF & 14.9 & 15.9 & 16.0 & 16.1 & 16.6 \\
  \hline
  \multirow{2}{*}{\shortstack[l]{ICL office kt0n}} & SoTA & 60.9 & 73.5 & 83.2 & 92.8 & 90.7 \\
  & DTSDF & 60.9 & 73.7 & 81.2 & 89.7 & 88.6 \\
  \hline
  \multirow{2}{*}{\shortstack[l]{ICL office kt2n}} & SoTA & 53.3 & 59.3 & 64.7 & 63.3 & 65.1 \\
  & DTSDF & -- & 59.4 & 63.9 & 65.4 & 65.6 \\
  \hline
  \multirow{2}{*}{\shortstack[l]{ICL office kt3n}} & SoTA & 218.3 & 153.6 & 85.1 & 104.6 & 95.3 \\
  & DTSDF & 40.0 & 252.4 & 121.8 & 141.8 & 145.3 \\
  \hline
  \multirow{2}{*}{\shortstack[l]{Zhou lr1}} & SoTA & 22.5 & 24.8 & 25.7 & 27.9 & 31.1 \\
  & DTSDF & 22.5 & 24.7 & 25.6 & 27.4 & 30.1 \\
  \hline
  \multirow{2}{*}{\shortstack[l]{Zhou lr2}} & SoTA & 17.5 & 19.0 & 21.6 & 25.0 & 28.5 \\
  & DTSDF & 17.3 & 18.6 & 20.9 & 23.9 & 27.2 \\
  \hline
  \multirow{2}{*}{\shortstack[l]{Zhou office1}} & SoTA & 29.0 & 32.7 & 36.4 & 47.2 & 54.1 \\
  & DTSDF & 29.1 & 32.5 & 35.0 & 38.2 & 43.3 \\
  \hline
  \multirow{2}{*}{\shortstack[l]{Zhou office2}} & SoTA & 25.9 & 29.0 & 34.3 & 84.3 & 64.6 \\
  & DTSDF & 25.8 & 28.1 & 31.8 & 37.1 & 44.6 \\
  \hline
  \multirow{2}{*}{\shortstack[l]{armadillo}} & SoTA & 3.6 & 4.5 & 10.9 & 20.4 & 31.4 \\
  & DTSDF & 3.5 & 4.1 & 6.8 & 12.1 & 19.2 \\
  \hline
  \multirow{2}{*}{\shortstack[l]{bunny}} & SoTA & 2.1 & 2.9 & 8.6 & 17.9 & 26.6 \\
  & DTSDF & 2.0 & 2.6 & 5.0 & 9.8 & 15.8 \\
  \hline
  \multirow{2}{*}{\shortstack[l]{dragon}} & SoTA & 4.0 & 5.8 & 15.2 & 25.2 & 33.2 \\
  & DTSDF & 3.9 & 4.9 & 9.4 & 16.5 & 25.3 \\
  \hline
  \multirow{2}{*}{\shortstack[l]{turbine blade}} & SoTA & 2.5 & 5.1 & 13.8 & 37.0 & 42.3 \\
  & DTSDF & 2.3 & 3.6 & 7.0 & 11.9 & 23.9 \\
  \hline
  \multirow{2}{*}{\shortstack[l]{TUM fr1 desk1}} & SoTA & 62.2 & 49.7 & 49.0 & 51.9 & 56.1 \\
  & DTSDF & 51.7 & 47.9 & 48.1 & 50.1 & 54.3 \\
  \hline
  \multirow{2}{*}{\shortstack[l]{TUM fr3 long\\office}} & SoTA & 43.4 & 43.0 & 44.7 & 51.2 & 61.3 \\
  & DTSDF & 37.3 & 38.1 & 40.8 & 47.4 & 55.4 \\
  \hline
  \multirow{2}{*}{\shortstack[l]{TUM fr3 structure\\notexture far}} & SoTA & 60.3 & 68.1 & 27.8 & 22.3 & 26.0 \\
  & DTSDF & 49.4 & 67.1 & 27.8 & 22.3 & 25.7 \\
  \hline
  \multirow{2}{*}{\shortstack[l]{TUM fr3 structure\\notexture near}} & SoTA & 7.0 & 7.5 & 7.7 & 8.1 & 8.9 \\
  & DTSDF & 6.4 & 6.7 & 6.9 & 64.3 & 7.8 \\
  \hline
  \multirow{2}{*}{\shortstack[l]{TUM fr3 structure\\texture far}} & SoTA & 41.6 & 26.2 & 22.1 & 24.0 & 25.7 \\
  & DTSDF & 40.4 & 25.4 & 21.6 & 22.9 & 25.1 \\
  \hline
  \multirow{2}{*}{\shortstack[l]{TUM fr3 structure\\texture near}} & SoTA & 46.9 & 69.5 & 103.7 & 31.0 & 28.2 \\
  & DTSDF & 43.5 & 38.8 & 126.8 & 93.8 & 25.8 \\
\end{tabular}
\end{table}

\bibliographystyle{IEEEtran}
\bibliography{references}

\begin{thebibliography}{10}
\providecommand{\url}[1]{#1}
\csname url@samestyle\endcsname
\providecommand{\newblock}{\relax}
\providecommand{\bibinfo}[2]{#2}
\providecommand{\BIBentrySTDinterwordspacing}{\spaceskip=0pt\relax}
\providecommand{\BIBentryALTinterwordstretchfactor}{4}
\providecommand{\BIBentryALTinterwordspacing}{\spaceskip=\fontdimen2\font plus
\BIBentryALTinterwordstretchfactor\fontdimen3\font minus
  \fontdimen4\font\relax}
\providecommand{\BIBforeignlanguage}[2]{{%
\expandafter\ifx\csname l@#1\endcsname\relax
\typeout{** WARNING: IEEEtran.bst: No hyphenation pattern has been}%
\typeout{** loaded for the language `#1'. Using the pattern for}%
\typeout{** the default language instead.}%
\else
\language=\csname l@#1\endcsname
\fi
#2}}
\providecommand{\BIBdecl}{\relax}
\BIBdecl

\bibitem{Newcombe2011}
R.~A. Newcombe, S.~Izadi, O.~Hilliges, D.~Molyneaux, D.~Kim, A.~J. Davison,
  P.~Kohi, J.~Shotton, S.~Hodges, and A.~Fitzgibbon, ``{KinectFusion}:
  Real-time dense surface mapping and tracking,'' in \emph{{IEEE} and {ACM}
  International Symposium on Mixed and Augmented Reality {(ISMAR)}}, 2011, pp.
  127--136.

\bibitem{Splietker2019}
M.~{Splietker} and S.~{Behnke}, ``Directional {TSDF}: Modeling surface
  orientation for coherent meshes,'' in \emph{IEEE/RSJ International Conference
  on Intelligent Robots and Systems (IROS)}, 2019, pp. 1727--1734.

\bibitem{Oleynikova2017}
H.~Oleynikova, Z.~Taylor, M.~Fehr, R.~Siegwart, and J.~Nieto, ``{Voxblox}:
  Incremental {3D} {E}uclidean signed distance fields for on-board {MAV}
  planning,'' in \emph{IEEE/RSJ International Conference on Intelligent Robots
  and Systems (IROS)}, Sep. 2017, pp. 1366--1373.

\bibitem{Mescheder2019a}
L.~Mescheder, M.~Oechsle, M.~Niemeyer, S.~Nowozin, and A.~Geiger, ``Occupancy
  networks: Learning {3D} reconstruction in function space,'' in \emph{IEEE
  Conference on Computer Vision and Pattern Recognition (CVPR)}, 2019, pp.
  4455--4465.

\bibitem{Park2019}
J.~J. Park, P.~Florence, J.~Straub, R.~Newcombe, and S.~Lovegrove, ``{DeepSDF}:
  Learning continuous signed distance functions for shape representation,'' in
  \emph{IEEE Conference on Computer Vision and Pattern Recognition (CVPR)},
  2019, pp. 165--174.

\bibitem{Mildenhall2020}
B.~Mildenhall, P.~P. Srinivasan, M.~Tancik, J.~T. Barron, R.~Ramamoorthi, and
  R.~Ng, ``{NeRF}: Representing scenes as neural radiance fields for view
  synthesis.''

\bibitem{Azinovic2021}
D.~Azinovi{\'c}, R.~Martin-Brualla, D.~B. Goldman, M.~Nie{\ss}ner, and
  J.~Thies, ``Neural {RGB-D} surface reconstruction,'' \emph{arXiv:2104.04532},
  2021.

\bibitem{Dong2018a}
W.~Dong, Q.~Wang, X.~Wang, and H.~Zha, ``{PSDF} {F}usion: Probabilistic signed
  distance function for on-the-fly {3D} data fusion and scene reconstruction,''
  in \emph{European Conference on Computer Vision (ECCV)}, 2018, pp. 701--717.

\bibitem{Henry2013}
P.~Henry, D.~Fox, A.~Bhowmik, and R.~Mongia, ``Patch volumes: Multiple fusion
  volumes for consistent {RGB-D} modeling,'' in \emph{RSS workshop on RGB-D:
  Advanced reasoning with depth cameras, Berlin, Germany}, 2013.

\bibitem{Millane2018}
A.~Millane, Z.~Taylor, H.~Oleynikova, J.~Nieto, R.~Siegwart, and C.~Cadena,
  ``C-blox: A scalable and consistent {TSDF}-based dense mapping approach,'' in
  \emph{IEEE/RSJ International Conference on Intelligent Robots and Systems
  (IROS)}, 2018, pp. 995--1002.

\bibitem{Zhang2021}
S.~Zhang, L.~Zheng, and W.~Tao, ``Survey and evaluation of {RGB-D} {SLAM},''
  \emph{IEEE Access}, vol.~9, pp. 21\,367--21\,387, 2021.

\bibitem{Prisacariu2017}
V.~A. Prisacariu, O.~K{\"a}hler, S.~Golodetz, M.~Sapienza, T.~Cavallari, P.~H.
  Torr, and D.~W. Murray, ``{InfiniTAM} v3: A framework for large-scale {3D}
  reconstruction with loop closure,'' \emph{arXiv:1708.00783}, 2017.

\bibitem{Kerl2013}
C.~Kerl, J.~Sturm, and D.~Cremers, ``Dense visual {SLAM} for {RGB-D} cameras,''
  in \emph{IEEE/RSJ International Conference on Intelligent Robots and Systems
  (IROS)}, 2013, pp. 2100--2106.

\bibitem{Whelan2015}
T.~Whelan, M.~Kaess, H.~Johannsson, M.~Fallon, J.~J. Leonard, and J.~McDonald,
  ``Real-time large-scale dense {RGB-D} {SLAM} with volumetric fusion,''
  \emph{The International Journal of Robotics Research}, vol.~34, no. 4-5, pp.
  598--626, 2015.

\bibitem{Dai2017}
A.~Dai, M.~Nie{\ss}ner, M.~Zollh{\"o}fer, S.~Izadi, and C.~Theobalt,
  ``{Bundlefusion}: Real-time globally consistent {3D} reconstruction using
  on-the-fly surface reintegration,'' \emph{ACM Transactions on Graphics
  (TOG)}, vol.~36, no.~3, p.~24, 2017.

\bibitem{Bylow2016}
E.~Bylow, C.~Olsson, and F.~Kahl, ``Robust online {3D} reconstruction combining
  a depth sensor and sparse feature points,'' in \emph{International Conference
  on Pattern Recognition (ICPR)}, 2016, pp. 3709--3714.

\bibitem{Canelhas2013}
D.~R. Canelhas, T.~Stoyanov, and A.~J. Lilienthal, ``Sdf tracker: A parallel
  algorithm for on-line pose estimation and scene reconstruction from depth
  images,'' in \emph{IEEE/RSJ International Conference on Intelligent Robots
  and Systems (IROS)}, 2013, pp. 3671--3676.

\bibitem{Slavcheva2018}
M.~Slavcheva, W.~Kehl, N.~Navab, and S.~Ilic, ``{SDF-2-SDF} registration for
  real-time {3D} reconstruction from {RGB-D} data,'' \emph{International
  Journal of Computer Vision}, vol. 126, pp. 615--636, 2018.

\bibitem{Millane2020}
A.~Millane, H.~Oleynikova, C.~Lanegger, J.~Delmerico, J.~Nieto, R.~Siegwart,
  M.~Pollefeys, and C.~Cadena, ``Freetures: Localization in signed distance
  function maps,'' \emph{arXiv:2010.09378}, 2020.

\bibitem{Bylow2013}
E.~Bylow, J.~Sturm, C.~Kerl, F.~Kahl, and D.~Cremers, ``Real-time camera
  tracking and {3D} reconstruction using signed distance functions.'' in
  \emph{Robotics: Science and Systems (RSS)}, vol.~2, 2013.

\bibitem{Nguyen2012}
C.~V. Nguyen, S.~Izadi, and D.~Lovell, ``Modeling {Kinect} sensor noise for
  improved {3D} reconstruction and tracking,'' in \emph{International
  Conference on 3D Imaging, Modeling, Processing, Visualization and
  Transmission (3DIMPVT)}, 2012, pp. 524--530.

\bibitem{Dryanovski2017}
I.~Dryanovski, M.~Klingensmith, S.~S. Srinivasa, and J.~Xiao, ``Large-scale,
  real-time {3D} scene reconstruction on a mobile device,'' \emph{Autonomous
  Robots}, vol.~41, pp. 1423--1445, 2017.

\bibitem{Dong2018}
W.~Dong, J.~Shi, W.~Tang, X.~Wang, and H.~Zha, ``An efficient volumetric mesh
  representation for real-time scene reconstruction using spatial hashing,''
  2018.

\bibitem{Klingensmith2015}
M.~Klingensmith, I.~Dryanovski, S.~Srinivasa, and J.~Xiao, ``Chisel: Real time
  large scale {3D} reconstruction onboard a mobile device using spatially
  hashed signed distance fields.'' in \emph{Robotics: Science and Systems
  (RSS)}, vol.~4, 2015.

\bibitem{Kaehler2015}
O.~K{\"a}hler, V.~Prisacariu, J.~Valentin, and D.~Murray, ``Hierarchical voxel
  block hashing for efficient integration of depth images,'' \emph{IEEE
  Robotics and Automation Letters}, vol.~1, no.~1, pp. 192--197, 2015.

\bibitem{Niesner2013}
M.~Nie{\ss}ner, M.~Zollh{\"o}fer, S.~Izadi, and M.~Stamminger, ``Real-time {3D}
  reconstruction at scale using voxel hashing,'' \emph{ACM Transactions on
  Graphics (ToG)}, vol.~32, no.~6, p. 169, 2013.

\bibitem{Stotko2019}
P.~Stotko, ``stdgpu: Efficient {STL}-like data structures on the {GPU},''
  \emph{arXiv:2010.09378}, 2019.

\bibitem{Zhou2013}
Q.-Y. Zhou and V.~Koltun, ``Dense scene reconstruction with points of
  interest,'' \emph{ACM Transactions on Graphics}, vol.~32, no.~4, p. 112,
  2013.

\bibitem{Handa2014}
A.~Handa, T.~Whelan, J.~McDonald, and A.~J. Davison, ``A benchmark for {RGB-D}
  visual odometry, {3D} reconstruction and {SLAM},'' in \emph{IEEE
  International Conference on Robotics and Automation (ICRA)}, 2014, pp.
  1524--1531.

\bibitem{Choi2015}
S.~Choi, Q.-Y. Zhou, and V.~Koltun, ``Robust reconstruction of indoor scenes,''
  in \emph{IEEE Conference on Computer Vision and Pattern Recognition (CVPR)},
  2015, pp. 5556--5565.

\bibitem{Sturm2012}
J.~Sturm, N.~Engelhard, F.~Endres, W.~Burgard, and D.~Cremers, ``A benchmark
  for the evaluation of {RGB-D} {SLAM} systems,'' in \emph{IEEE/RSJ
  International Conference on Intelligent Robots and Systems (IROS)}, 2012, pp.
  573--580.

\end{thebibliography}

\end{document}